\documentclass[runningheads]{llncs}

\usepackage{times}
\usepackage{latexsym}
\usepackage{empheq}
\usepackage{wrapfig}
\usepackage[T1]{fontenc}
\usepackage[utf8]{inputenc}

\usepackage{microtype}
\usepackage[most]{tcolorbox}
\usepackage{tabularx}
\usepackage{inconsolata}

\usepackage{graphicx}

\usepackage{amsfonts}       % blackboard math symbols
\usepackage{amsmath}
\usepackage{booktabs}       % professional-quality tables
\usepackage{booktabs}
\usepackage{xcolor}
\usepackage{colortbl}
\usepackage{multirow}
\usepackage{siunitx}   % for S column type (decimal alignment)
\definecolor{headerblue}{RGB}{30,  90, 160}   % deep blue  — header rows
\definecolor{rowgray}   {RGB}{240, 243, 247}  % light slate — alternating data rows
\definecolor{bestgold}  {RGB}{255, 243, 205}  % soft gold   — best result rows
\definecolor{hdrtext}   {RGB}{255, 255, 255}  % white       — header text
\makeatletter
\renewcommand\section{\@startsection{section}{1}{\z@}%
                       {-8\p@ \@plus -4\p@ \@minus -2\p@}% % Spacing before (Default is usually larger)
                       {4\p@ \@plus 2\p@ \@minus 1\p@}%   % Spacing after
                       {\normalfont\large\bfseries\rightskip=\z@ \@plus2em\leftskip=\z@}}

\renewcommand\subsection{\@startsection{subsection}{2}{\z@}%
                       {-6\p@ \@plus -3\p@ \@minus -1\p@}% % Spacing before
                       {3\p@ \@plus 1\p@ \@minus 1\p@}%   % Spacing after
                       {\normalfont\normalsize\bfseries\rightskip=\z@ \@plus2em\leftskip=\z@}}
\makeatother

\title{SelfGraphRAG: Bridging the Supervision Gap in Graph-Based RAG with Synthetic QA Generation}
\titlerunning{SelfGraphRAG}
\author{
Ben Lagnese \and Manas Gaur
}

\authorrunning{B. Lagnese and M. Gaur}

\institute{
University of Maryland, Baltimore County (UMBC), Baltimore, MD, USA \\
\email{\{blagnese, mgaur\}@umbc.edu} \\
\url{https://kai2.umbc.edu}
}

\begin{document}
\maketitle
\begin{abstract}
Retrieval-augmented generation (RAG) enhances large language models by incorporating external knowledge without retraining, yet existing approaches fail to fully exploit relational structure in knowledge graphs. While graph-based RAG captures entity relationships, its effectiveness is limited by the lack of labeled data needed for supervised graph retrieval. We study whether synthetic supervision derived from the graph itself can enable effective learning of query-conditioned subgraph retrieval. We propose SelfGraphRAG, a framework that generates question–answer pairs from knowledge graph structures to induce training signals for a graph retrieval model. We hypothesize that multi-hop and neighborhood-based synthetic queries encode sufficient relational constraints to approximate real task distributions. Empirical results across multi-hop and classification benchmarks demonstrate that models trained with such synthetic supervision improve retrieval precision and reasoning performance over embedding-based baselines. These findings suggest that graph-structured data can serve as a self-contained source of supervision, enabling scalable learning of structured retrieval without manual annotation.
\end{abstract}

\section{Introduction}

Large language models (LLMs) have demonstrated remarkable language understanding and generation capabilities, yet they remain fundamentally limited by their parameterized
knowledge; they cannot efficiently incorporate new information, are prone to hallucination, and have fixed context windows that cannot accommodate large private corpora \cite{rag}. Retrieval-augmented generation (RAG) addresses these limitations by indexing an external knowledge base and supplying relevant context at inference time, improving both accuracy and transparency \cite{rag}\cite{lightrag}. Standard RAG, however, operates on flat document chunks. This representation is ill-suited for questions that require reasoning across entity relationships, precisely the queries that arise in complex, knowledge-intensive domains. A natural remedy is to extract a knowledge graph from the corpus during indexing, making entity relationships explicitly retrievable \cite{graphrag}\cite{lightrag}. Entity-level strategies retrieve local neighborhoods of topic nodes and pass the surrounding edges as context, while corpus-level strategies partition the graph into communities, summarize each community with an LLM, and produce global answers. Both approaches, however, are limited: entity-level retrieval relies on embedding similarity with fixed hop sizes, risking both redundancy and incomplete multi-hop
coverage; corpus-level summarization sacrifices the granularity of individual triples, degrading answer precision.

Graph retrieval models, which learn to score and extract task-relevant subgraphs from a knowledge graph, can overcome these limitations, but they introduce a new requirement: a question-answer dataset labeled against the target graph for supervised training \cite{gretriever}\cite{subgraphrag}. When an LLM constructs the graph from private documents, no such labeled data exists. This creates a fundamental gap between the expressiveness of graph retrieval models and their practical applicability to knowledge-graph-augmented RAG pipelines.

We propose \textbf{SelfGraphRAG}, a pipeline that closes this gap through synthetic data generation. Given an automatically extracted knowledge graph, SelfGraphRAG prompts an LLM
to construct two complementary types of question-answer pairs: multi-hop questions that require reasoning across non-adjacent nodes, and node-summarization questions that require integrating evidence from a node's immediate neighborhood, to produce a labeled training set. We then train a graph retrieval model on this synthetic corpus and use it for inference, replacing the embedding-based similarity retrieval used in standard graph RAG methods. SelfGraphRAG's contributions are threefold. First, we demonstrate that synthetic QA generation from extracted knowledge graphs produces a training signal sufficient to outperform both standard RAG and LightRAG on multi-hop and classification benchmarks. Second, we show that the inference-time gain is achieved with a compact, locally runnable model stack, without reliance on proprietary APIs. Third, we provide a reusable pipeline that can be applied to any domain with unlabeled document corpora and prohibitively expensive annotation.

% related work
\section{Related Work}

\subsection{Retrieval-Augmented Generation}

Retrieval-augmented generation (RAG) augments LLM outputs by retrieving relevant passages
from an indexed external corpus and prepending them as context at inference time
\cite{rag}. This approach directly mitigates key failure modes of parametric LLMs,
including hallucination, stale knowledge, and the inability to accommodate private corpora, without requiring costly retraining \cite{rag,lightrag,tilwani2024neurosymbolic}. 

Standard RAG systems operate
in three stages: offline indexing of chunked documents into a vector store, query-time
retrieval of relevant chunks via embedding similarity, and context-conditioned generation \cite{saxena2025ranking}.
While effective for surface-level factual recall, flat chunk retrieval cannot support
reasoning over entity relationships or corpus-wide synthesis, since inter-entity dependencies
are fragmented across independently indexed chunks \cite{lightrag,graphrag,tilwani2024reasons,guttal2026structure,saxena2026neurosymbolic}.
\textit{SelfGraphRAG directly addresses this limitation}: by replacing chunk-level retrieval with a graph retrieval model trained on synthetically generated supervision, it preserves the core RAG architecture while enabling multi-hop relational reasoning that flat-retrieval systems fundamentally cannot support, a natural extension to ISEEQ \cite{gaur2022iseeq}.

\subsection{Knowledge Graph Extraction and Graph-Based RAG}

To address the relational limitations of standard RAG, a family of methods constructs a
knowledge graph from the document corpus during indexing and uses the graph structure to
guide retrieval \cite{graphrag,lightrag}. GraphRAG supports
both entity-level retrieval, retrieving local neighborhoods of topic nodes, and
corpus-level retrieval via community detection and LLM-based summarization of each
community, enabling global sensemaking over large corpora. LightRAG
\cite{lightrag} simplifies this pipeline with dual-level retrieval but similarly relies on
embedding similarity for entity matching. Both systems inherit a fundamental limitation:
embedding-based retrieval assigns relevance based on surface-level semantic similarity to
the query, which fails for multi-hop questions whose relevant nodes may be semantically
distant from the query string \cite{subgraphrag}. \textit{SelfGraphRAG builds directly on
the Doc2Graph stage of GraphRAG}, reusing its entity and relation extraction pipeline
without modification. It then departs from both GraphRAG and LightRAG by replacing
embedding-similarity retrieval with a supervised graph retrieval model, targeting precisely
the multi-hop failure mode that neither summarization nor embedding-based approaches can
resolve.

A complementary line of work proposes recursive abstractive summarization over retrieved
text hierarchies. RAPTOR \cite{raptor} constructs a tree of progressively abstracted summaries and retrieves at multiple levels of granularity, improving coverage of
corpus-level questions. Unlike community summarization, RAPTOR preserves original passage
text at leaf nodes, partially mitigating granularity loss. However, neither RAPTOR nor
summarization-based GraphRAG approaches learn to retrieve based on question structure; their retrieval quality is bounded by embedding similarity regardless of the complexity
of the query. SelfGraphRAG works in the direction of RAPTOR's motivation, retaining
fine-grained evidence while supporting broad query coverage, but achieves this through
trained graph retrieval rather than hierarchical summarization, avoiding the information
loss that accompanies any abstractive step.

HippoRAG \cite{hipporag} takes a structurally motivated alternative by modeling graph
retrieval as personalized PageRank over the extracted knowledge graph, simulating
associative memory recall without any training. While training-free, HippoRAG's retrieval
quality is directly bound by the quality of the extracted graph structure and cannot be
improved through labeled supervision or task-specific fine-tuning. \textit{SelfGraphRAG improves
upon HippoRAG's core intuition} that graph traversal should drive retrieval, but replaces
the fixed PageRank heuristic with a learned scoring function. This allows SelfGraphRAG to
adapt retrieval behavior to the question type, multi-hop inference versus neighborhood
summarization, in a way that a training-free traversal strategy cannot.

\subsection{Graph Retrieval Models}

Graph retrieval models learn to identify question-relevant subgraphs from a knowledge
graph, replacing heuristic traversal with trained scoring \cite{gretriever}\cite{subgraphrag}.
G-Retriever \cite{gretriever} combines a graph transformer with a frozen LLM,
training end-to-end on graph question-answer datasets to score and retrieve question-relevant
subgraphs; it is the GraphLM backbone employed in SelfGraphRAG. \textit{SelfGraphRAG directly builds
upon G-Retriever}, preserving its architecture while solving the core barrier to its
application: the absence of labeled QA data for privately extracted knowledge graphs.
Rather than requiring hand-annotated data, SelfGraphRAG generates the training set that
G-Retriever needs from the extracted graph itself, making the model applicable to any
unlabeled corpus without additional annotation effort.

\textbf{SubgraphRAG} \cite{subgraphrag} extends the graph retrieval paradigm with a
lightweight MLP-based triple scorer and a parallel triple-scoring mechanism for efficient,
flexible subgraph retrieval, but it requires both labeled QA pairs \emph{and} topic entity
annotations to initialize retrieval. GNN-RAG \cite{gnnrag} trains a GNN to score
answer candidate nodes and extracts shortest-path reasoning chains for LLM consumption,
achieving state-of-the-art on WebQSP and CWQ; it similarly requires pre-specified question
entities and labeled QA data tied to the target graph. \textit{SelfGraphRAG works in the
direction of both SubgraphRAG and GNN-RAG}, learning to retrieve structurally
meaningful subgraphs rather than relying on embedding similarity, but removes their
shared dependency on external supervision. The synthetic QA pairs generated by SelfGraphRAG's
SynthGen stage provide the labeled signal that SubgraphRAG and GNN-RAG assume is given,
extending the applicability of supervised graph retrieval to the unlabeled, self-constructed
graph setting \cite{tilwani2026neurosymbolic}.

The critical barrier shared by all graph retrieval models is the need for a labeled
question-answer dataset aligned to the target knowledge graph\cite{gaur2022iseeq}. When the graph is
constructed by automated extraction from private documents, as in GraphRAG and LightRAG, no such dataset exists. SynthKG \cite{synthkg} independently proposes
synthetic data generation for the complementary task of training the \emph{graph extractor}
rather than the graph retriever, demonstrating that LLM-generated supervision can match or
exceed the quality of human annotation for structured extraction tasks. \textit{SelfGraphRAG
applies the same self-supervised philosophy to the retrieval stage}, treating SynthKG as a
proof of concept that synthetic graph-derived data can substitute for human annotation, and
extending it to produce training signal for the downstream graph retrieval model rather than
for graph construction.

% problem definition
\section{SelfGraphRAG}

\begin{figure*}[t]
    \centering
    \includegraphics[width=\textwidth]{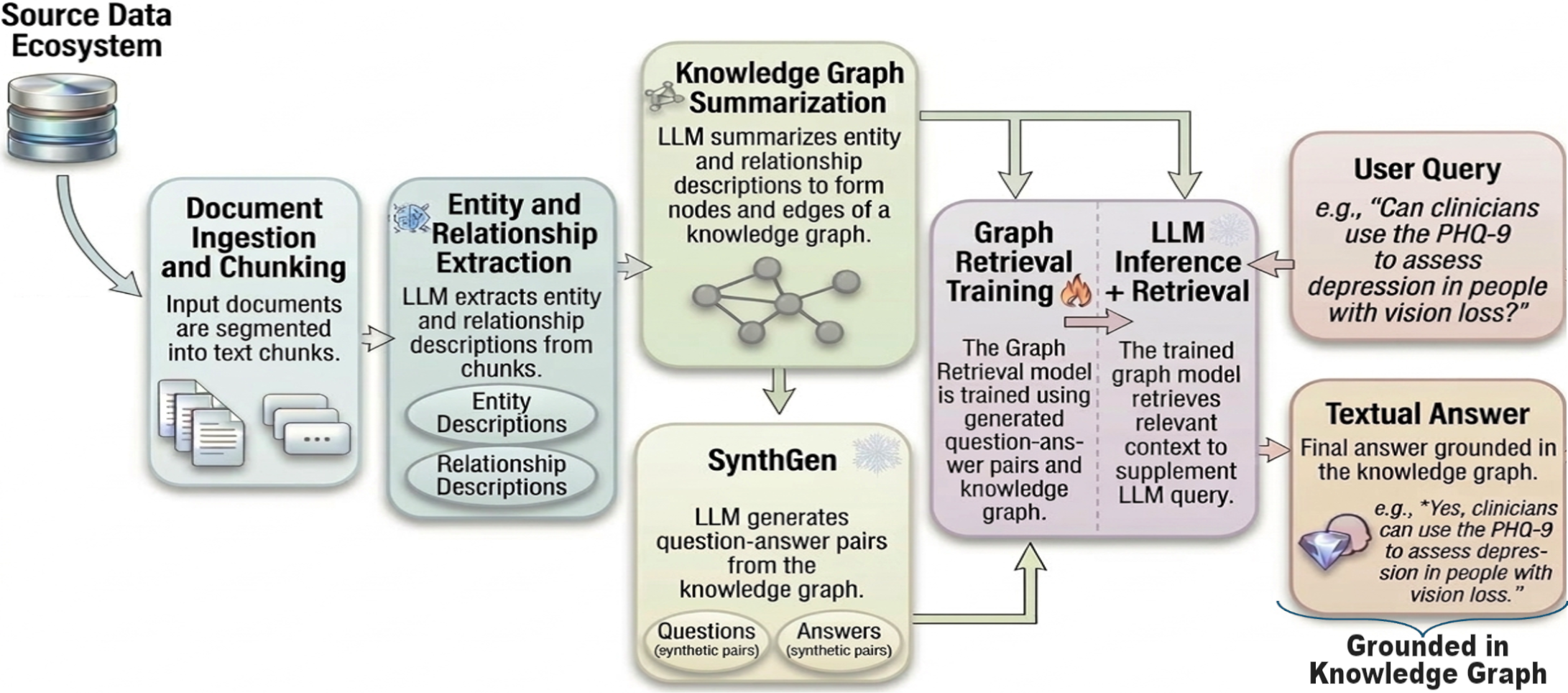}
    \caption{The SelfGraphRAG process. Indexing mirrors GraphRAG through knowledge graph
summarization. SynthGen then generates a question-answer dataset to train a graph
retrieval model, which retrieves information from the graph at inference time.}
    \label{fig:selfgraphrag_diagram}
\end{figure*}

\noindent \textbf{Task Definition:} We consider the open-domain question answering task over a private, unlabeled document corpus. This setting is the deployment scenario
motivating SelfGraphRAG: an organization holds a set of internal documents - clinical notes, technical reports, or similar, and wants a system that can answer natural
language questions against that corpus, but has no pre-existing question-answer pairs tied to it and no practical way to produce them by hand at the scale required to train a supervised retriever. 

Formally, let $\mathcal{D} = \{D_1, D_2, \dots, D_d\}$ denote a collection of $d$ text documents. Given a natural language query $q$, the goal is to produce a natural language answer $a$ intended to be grounded in $\mathcal{D}$, in the sense that the content of $a$ should be traceable to facts stated somewhere in $\mathcal{D}$ rather than drawn from the answering model's own parametric knowledge. We assume $\mathcal{D}$ is unlabeled: no question-answer pairs annotated against
$\mathcal{D}$ are assumed to exist prior to indexing, and no ground-truth supervision is provided to any system at index time. This assumption is what rules out directly training a supervised graph retriever on $\mathcal{D}$ and is the premise the rest of this section formalizes.

The core challenge SelfGraphRAG addresses is a mismatch between what the best-performing retrieval strategy needs and what the unlabeled setting provides, graph retrieval models learn to score question-relevant subgraphs,
but training them requires a labeled question-answer dataset grounded in the target graph, and by assumption no such dataset exists for $\mathcal{D}$. One way to avoid this requirement altogether is to fall back on embedding-based retrieval, which needs no labeled data - but, as discussed in Section~2, embedding-based retrieval systematically under-serves multi-hop questions whose answer-relevant nodes are not lexically similar to the query, which is precisely the failure mode graph retrieval is meant to correct. SelfGraphRAG does not take this route. Instead, it self-supervises: rather than relying on an external labeled dataset, it uses the corpus graph's own structure to determine what to ask, and an auxiliary LLM call to generate the question-answer pairs themselves, so that a supervised retriever becomes trainable even though no human-annotated data exists for $\mathcal{D}$.

\noindent \textbf{Problem Formulation:} Let $M(\cdot\,;\, \pi)$ denote a call to a large
language model conditioned on system prompt $\pi$, taking a text input and returning a
text output. We use this notation uniformly for all LLM invocations throughout the
pipeline, with the prompt $\pi$ distinguishing each distinct role (extraction,
summarization, question generation); the notation is deliberately role-agnostic because
every stage of SelfGraphRAG - indexing, synthetic data generation, and answer generation
- is implemented as a differently-prompted call to the same class of underlying model.

Let $\mathcal{G} = (V, E)$ denote a knowledge graph extracted from $\mathcal{D}$. Each
node $v \in V$ is a textual description of an entity mentioned in the corpus (for
example, a clinical procedure, a person, or a condition), and each directed edge
$e = (v_i,\, v_j,\, r) \in E$ encodes a labeled relation $r$ from entity $v_i$ to entity
$v_j$, with $v_i, v_j \in V$. The graph is thus a structured, queryable proxy for the
relational content of $\mathcal{D}$: information that in the source documents is
scattered across sentences and chunks becomes, after extraction, a set of explicit
entity-to-entity edges that a retriever can traverse. We write $e_{ij}$ as shorthand for
the edge $(v_i, v_j, r) \in E$ directed from $v_i$ to $v_j$; when the relation label $r$
is needed explicitly, we write $r(v_i, v_j)$.

For any subgraph $\hat{\mathcal{G}} = (\hat{V}, \hat{E}) \subseteq \mathcal{G}$ - the hat
distinguishing a subgraph selected out of $\mathcal{G}$ from the full graph itself - let
$\mathrm{Verbalize}(\hat{\mathcal{G}})$ denote a linearization of $\hat{\mathcal{G}}$ into a natural language string, formed by converting each triple $(v_i, v_j, r) \in \hat{E}$ into the sentence ``$v_i$ \texttt{[r]} $v_j$'' and concatenating the resulting sentences in edge-insertion order. This procedure is deterministic: the sentence template and the edge-insertion ordering are both fixed, so the same subgraph always maps to the same string, with no sampling or model call involved. We define $\mathrm{Verbalize}$
generically over any subgraph because it is applied twice later in this section to two
different kinds of subgraph: once to the \emph{retrieved} subgraph $\hat{\mathcal{G}}_q$
that a trained retriever returns for a query $q$ (Equation~\ref{eq:answer}, below), and
once to the small \emph{induced} subgraphs used to construct synthetic training questions
in Stage 2. Determinism matters here: it lets us attribute any variability in downstream
answers to which subgraph was retrieved, rather than to inconsistency in how the subgraph
was described.

\paragraph{Supervised graph retrieval.}
Let $f_\theta : (q,\, \mathcal{G}) \to \hat{\mathcal{G}}_q$ denote a parameterized graph
retrieval model with learnable parameters $\theta$ that, given query $q$ and graph
$\mathcal{G}$, returns a retrieved subgraph $\hat{\mathcal{G}}_q \subseteq \mathcal{G}$.
Unlike embedding-based retrieval, which scores nodes independently by similarity to $q$,
$f_\theta$ is trained to score entire substructures of $\mathcal{G}$ jointly, which is
what in principle allows it to recover relevant nodes that are several hops from any
term in $q$. Let $\mathrm{LM}$ denote a large language model with \emph{frozen}
parameters used solely for answer generation, distinct from the extraction and
generation LLMs used during indexing; freezing $\mathrm{LM}$ isolates the object being
learned to $f_\theta$ alone, so that any change in answer quality across training runs
can be attributed to the retriever rather than to drift in the answering model. The
answer to query $q$ is produced as:
\begin{equation*}
    a \;=\; \mathrm{LM}\!\left(q,\;\mathrm{Verbalize}(\hat{\mathcal{G}}_q)\right).
    \label{eq:answer}
\end{equation*}

Training $f_\theta$ requires a dataset $\mathcal{T} = \{(q_i,\, a_i)\}_{i=1}^{N}$ of $N$
question-answer pairs grounded in $\mathcal{G}$: without such pairs, there is no signal
by which $f_\theta$ can learn to distinguish a relevant subgraph from an irrelevant one.
Since $\mathcal{D}$ is unlabeled, $\mathcal{T}$ is unavailable. SelfGraphRAG synthesizes
$\mathcal{T}$ directly from $\mathcal{G}$, replacing manual annotation with structured
LLM-driven question generation over graph substructures - the mechanism detailed in the
SynthGen stage below.

\smallskip
\noindent \textbf{Three Stages of SelfGraphRAG:} SelfGraphRAG operates in three sequential stages: \textit{Doc2Graph}, which constructs
$\mathcal{G}$ from $\mathcal{D}$; \textit{SynthGen}, which generates the synthetic training
set $\mathcal{T}$ from $\mathcal{G}$; and \textit{GraphLM}, which trains $f_\theta$ on
$\mathcal{T}$ and deploys it for inference. Figure \ref{fig:selfgraphrag_diagram} shows a high level overview of the steps involved.

%-------------------------------------------------------------------
\paragraph{Stage 1: Doc2Graph.}

Let $L \in \mathbb{Z}^+$ denote the maximum number of tokens per chunk. For each document
$D_i \in \mathcal{D}$, the chunking function
\[
    \mathrm{ChunkText}(D_i,\, L)
    \;=\; [C_{i,1},\; C_{i,2},\; \dots,\; C_{i,k_i}]
\]
partitions $D_i$ into $k_i \geq 1$ contiguous, non-overlapping text segments, where each
chunk $C_{i,j}$ satisfies $\mathrm{len}(C_{i,j}) \leq L$ tokens. The global chunk set is:
\[
    \mathcal{C} \;=\; \bigcup_{i=1}^{d}\; \mathrm{ChunkText}(D_i,\, L).
\]
Entity and relation extraction proceeds in two explicitly separate steps per chunk.
Let $\pi_\mathrm{ext}$ be the extraction prompt and $\pi_\mathrm{summ}$ be the
summarization prompt. For each chunk
$C \in \mathcal{C}$:

\textit{Step 1: Extraction.} The LLM identifies raw entity mentions
$\mathcal{E}_C = \{\varepsilon_1, \dots, \varepsilon_p\}$ and raw relation mentions
$\mathcal{R}_C = \{\rho_1, \dots, \rho_q\}$ from $C$:
$
    (\mathcal{E}_C,\, \mathcal{R}_C)
    \;=\; M(C\,;\, \pi_\mathrm{ext}).
$

\textit{Step 2: Summarization.} Each raw mention is independently condensed into a
canonical node or edge label. $\pi_\mathrm{summ}$ is parameterized by a
\texttt{mention\_type} field (\texttt{entity} or \texttt{relation}), so that entity and
relation mentions receive type-appropriate canonicalization instructions despite sharing
one prompt symbol. The chunk-level node and edge sets are:
\begin{equation*}
    V_C = \bigl\{\, M(\varepsilon\,;\, \pi_\mathrm{summ})
             \;\mid\; \varepsilon \in \mathcal{E}_C \,\bigr\},
    E_C = \bigl\{\, M(\rho\,;\, \pi_\mathrm{summ})
             \;\mid\; \rho \in \mathcal{R}_C \,\bigr\}.
\end{equation*}
The global graph is the union over all chunks:
\[
    \mathcal{G} \;=\; \left(\bigcup_{C \in \mathcal{C}} V_C,\;\;
                            \bigcup_{C \in \mathcal{C}} E_C\right).
\]
This stage reuses the GraphRAG indexing pipeline \cite{graphrag} without modification.

The union above merges chunk-level node and edge sets without a deduplication or
coreference step. When the same real-world entity is mentioned in two different chunks and
independently canonicalized by $\pi_\mathrm{summ}$ into two different node labels,
$\mathcal{G}$ retains two disconnected nodes for one entity, and any relation that should
connect them across chunks is not captured. We inherit this design directly from the
unmodified GraphRAG indexing pipeline \cite{graphrag}; LightRAG addresses the same gap
with an explicit de-duplication step, which is one reason we treat entity resolution as an
open direction for strengthening $\mathcal{G}$'s coverage rather than a solved part of
Doc2Graph.

%-------------------------------------------------------------------
\paragraph{Stage 2: SynthGen.}

\begin{figure}[t]
    \centering
    \includegraphics[width=0.7\textwidth]{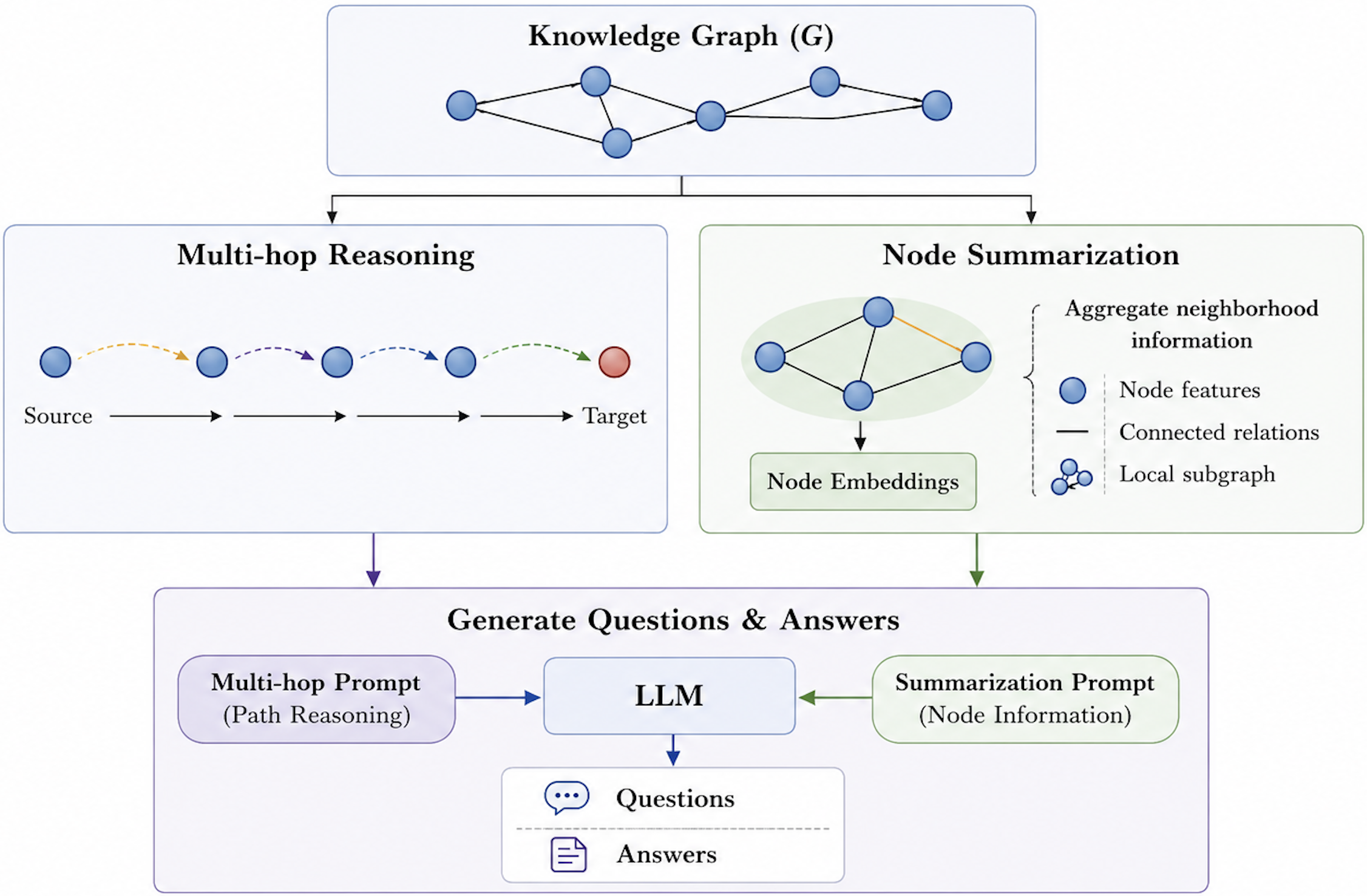}
    \caption{The SynthGen process. From the knowledge graph, two sets are formed: one of two-hop-connected nodes and another of 3 edges connected by singular nodes. The LLM is prompted to generate question-answer pairs from these sets, forming the SynthGen dataset.}
    \label{fig:synthgen_diagram}
\end{figure}

SynthGen generates two classes of question-answer pairs from $\mathcal{G}$ using
generation prompts $\pi_Q^\mathrm{hop}$ and $\pi_Q^\mathrm{sum}$ (full templates in
Appendix~\ref{app:prompts}), targeting distinct retrieval skills. Both prompt templates
are fixed before training and applied uniformly across the graph.

\textit{Multi-hop questions.} These pairs train $f_\theta$ to retrieve non-adjacent nodes
connected through intermediate entities. For each two-hop path in $\mathcal{G}$ - that
is, for each tuple $(v_p, v_o, v_s)$ such that $e_{po} \in E$ and $e_{os} \in E$, where
$v_p, v_o, v_s \in V$ are distinct and $e_{po} = (v_p, v_o, r_1)$,
$e_{os} = (v_o, v_s, r_2)$ for some relations $r_1, r_2$, the LLM generates a
question about the relationship between the endpoint nodes $v_p$ and $v_s$ mediated by
$v_o$: $(q^\mathrm{hop},\; a^\mathrm{hop})$
\[
    \;=\; M\!\left(\,
        \mathrm{Verbalize}(\{v_p, e_{po}, v_o, e_{os}, v_s\})
    \,;\, \pi_Q^\mathrm{hop}\right),
\]
where $\mathrm{Verbalize}$ is applied to the induced two-hop subgraph as defined above.
We enumerate two-hop paths as they occur in $\mathcal{G}$ rather than sampling among
them; in densely connected regions of $\mathcal{G}$ this can produce a large number of
candidate paths, and we do not apply an additional filtering step beyond the tuple
constraints given here.

We restrict $\mathcal{T}^\mathrm{hop}$ to two-hop paths, following established efforts in
knowledge-graph question answering that frame reasoning chains within a two-hop scope \cite{yih2016value},\cite{talmor2018web}, \cite{gnnrag}. Bounding path length in this way keeps
two-hop enumeration tractable regardless of the branching factor of $\mathcal{G}$. This is a hyperparameter of SynthGen rather than a fixed constraint of the method, and can be
extended to longer paths to capture relational chains beyond two hops.

\textit{Node-summarization questions.} These pairs train $f_\theta$ to aggregate evidence
from a target node's immediate neighborhood. For each node $v_p \in V$ and a sampled set
of three distinct in-neighbors $\{v_o, v_s, v_r\} \subseteq V$ such that
$e_{op} = (v_o, v_p, r_o) \in E$, $e_{sp} = (v_s, v_p, r_s) \in E$, and
$e_{rp} = (v_r, v_p, r_r) \in E$ for some relations $r_o, r_s, r_r$, the LLM generates
a question about $v_p$ using only the information carried by its incoming edges: $(q^\mathrm{sum},\; a^\mathrm{sum})$
$
    =\; M\!\left(\,
        \mathrm{Verbalize}(\{e_{op}, e_{sp}, e_{rp}\})
    \,;\, \pi_Q^\mathrm{sum}\right).$
Nodes with fewer than three in-neighbors do not admit a sample of the required size and
are excluded from $\mathcal{T}^\mathrm{sum}$. We fix the in-neighbor sample size to three, following established efforts that bound
neighborhood aggregation to a fixed size for tractable, well-formed context construction
\cite{hamilton2017graphsage}. Fixing the size in this way keeps the verbalized
neighborhood short enough for the LLM to reason over reliably, while still supplying more
than a single edge of supporting evidence. This is a hyperparameter of SynthGen rather
than a fixed constraint of the method, and can be increased to supply richer neighborhood
context for nodes whose local structure warrants it. The complete synthetic training dataset as seen in Figure \ref{fig:synthgen_diagram} is the union of both types:
$
    \mathcal{T}
    \;=\; \mathcal{T}^\mathrm{hop} \;\cup\; \mathcal{T}^\mathrm{sum},
    \qquad
    |\mathcal{T}| = N,
$
where $\mathcal{T}^\mathrm{hop} = \{(q_i^\mathrm{hop}, a_i^\mathrm{hop})\}$ and
$\mathcal{T}^\mathrm{sum} = \{(q_j^\mathrm{sum}, a_j^\mathrm{sum})\}$ collect all
generated pairs of each type, and $N = |\mathcal{T}^\mathrm{hop}| +
|\mathcal{T}^\mathrm{sum}|$ is the total number of training examples.

%-------------------------------------------------------------------
\paragraph{Stage 3: GraphLM.}

The graph retrieval model $f_\theta$ is trained on $\mathcal{T}$ with the frozen language
model $\mathrm{LM}$ held fixed throughout. Formally, let
$\ell : \mathcal{Y} \times \mathcal{Y} \to \mathbb{R}_{\geq 0}$ denote the token-level
cross-entropy loss over the vocabulary $\mathcal{Y}$, defined for a predicted token
distribution $\hat{y}$ and gold token sequence $y$ as
$\ell(\hat{y}, y) = -\sum_t \log P_{\hat{y}}(y_t)$. The training objective is:
\begin{equation*}
\begin{split}
    \theta^*
    \;=\; \arg\min_{\theta}\;
    \mathbb{E}_{(q,\, a)\,\sim\,\mathcal{T}}\!\left[\,
        \ell~\!\Bigl(
            \mathrm{LM}\bigl(q,\,
                \mathrm{Verbalize}(f_\theta(q, \mathcal{G}))\bigr),\;
            a
        \Bigr)
    \right],
\end{split}
\end{equation*}
where gradients are propagated through $f_\theta$ only; the parameters of $\mathrm{LM}$
receive no gradient updates -- a design choice that, as discussed in the faithfulness
remark of Section~7, decouples whether $a$ matches the gold answer $a_i$ from whether $a$
is actually grounded in the retrieved subgraph. At inference time, the trained retrieval
model supplies a query-specific subgraph to the same frozen $\mathrm{LM}$:
$
\hat{a}
    \;=\; \mathrm{LM}\!\left(
        q,\;\mathrm{Verbalize}\!\left(f_{\theta^*}(q, \mathcal{G})\right)
    \right).
$
We instantiate $f_\theta$ as G-Retriever \cite{gretriever}, a graph transformer trained
end-to-end, and $\mathrm{LM}$ as a frozen Llama2-7b model, matching the backbone used
across the RAG, GraphRAG, and LightRAG baselines in Section~4.2 so that retrieval method
remains the primary variable of comparison.
\begin{tcolorbox}[
    enhanced,
    breakable,
    colback=green!3,
    colframe=green!45!black,
    boxrule=0.6pt,
    arc=2pt,
    left=2pt, right=2pt, top=2pt, bottom=2pt, boxsep=1pt, % Tightened margins
    title={\textbf{MoreHopQA: Retrieval $\rightarrow$ Reasoning $\rightarrow$ Answer}},
    fonttitle=\small\bfseries
]
\small

\textbf{Question.} What is the square of the sum of the digits of the birth year of the person who wrote and illustrated a Japanese manga series based on a 16-year-old high school student, Ichitaka Seto?

\smallskip
\textbf{Retrieved relational evidence.} $\text{Ichitaka Seto} \rightarrow \text{\textit{I''s}} \rightarrow \text{Masakazu Katsura} \rightarrow \text{1962}$

\smallskip
\textbf{SelfGraphRAG Justification.} \textit{I''s} is written and illustrated by Masakazu Katsura, whose birth year is 1962. The sum of its digits is $1+9+6+2=18$; therefore $18^2=324$.

\smallskip
\textbf{Prediction:} \textbf{324} \quad \textbf{Gold:} \textbf{324} \hfill \textbf{\textcolor{green!45!black}{Correct}}
\end{tcolorbox}

This formulation trains $f_\theta$ on the union $\mathcal{T}^\mathrm{hop} \cup
\mathcal{T}^\mathrm{sum}$ without isolating the contribution of either synthetic QA type,
or of the G-Retriever architecture choice, to downstream performance.

% experimental setup
\section{Experimental Setup}

\subsection{Datasets}

We evaluate on three datasets selected to stress-test different dimensions of graph-based multi-hop retrieval. \textbf{MoreHopQA}  is a multi-hop question answering benchmark constructed to require reasoning across multiple documents and entity relationships, with diverse answer types including yes/no, numerical, date, and named-entity responses \cite{morehopqa}. It directly exercises the multi-hop retrieval capability that SelfGraphRAG's SynthGen component is designed to train. \textbf{MultiHop-RAG}  is a benchmark designed specifically to evaluate RAG systems on questions that require synthesizing information from multiple retrieved passages \cite{multihoprag}. Its constrained answer format (yes/no or named entity) makes token-based precision and recall well-defined, enabling crisp comparison across retrieval strategies. \textbf{PubMedQA} is a biomedical question answering dataset requiring yes/no/maybe classification over research abstracts \cite{pubmedqa}. Its inclusion tests whether SelfGraphRAG's graph-retrieval approach generalizes to domain-specific, classification-framed queries. For all three datasets, each row of source data is written to its own text document (or grouped into documents of 5-10 contexts for datasets with short, fragmented entries) to simulate the unlabeled fragmented corpora that SelfGraphRAG is designed for. Ground-truth
question-answer pairs from each dataset are used exclusively for evaluation.  SynthGen draws on the same per-benchmark document set described above. For each of
MoreHopQA, MultiHop-RAG, and PubMedQA, $\mathcal{T}$ is generated from the identical
$\mathcal{G}$ used at evaluation time, with no separate or held-out corpus. This matches
the deployment scenario in Section~3, where only one private, unlabeled corpus exists to
draw both training signal and evaluation queries from.

\subsection{Baselines}
We compare SelfGraphRAG against four systems spanning four points in the retrieval design space: flat chunk retrieval, corpus-level community summarization, entity-level embedding retrieval, and trained graph retrieval.

\textbf{RAG} (RoBERTa-Large embedder, Llama2-7b generator) is the flat-retrieval baseline, chunking and embedding documents with RoBERTa-Large so that the top-$k$ chunks can be retrieved by cosine similarity and passed to Llama2-7b for generation. Run directly through PyTorch, this baseline establishes the performance floor achievable without any graph structure. \textbf{GraphRAG} (Mistral-7b generator, Snowflake-arctic-embed2-568m embedder)
implements the community-detection and LLM-summarization pipeline of \cite{graphrag}, run through a local Ollama API. As SelfGraphRAG's immediate predecessor, it shares the Doc2Graph stage and instantiates the corpus-level summarization paradigm that SelfGraphRAG replaces with a trained retriever. \textbf{LightRAG} \cite{lightrag} (Mistral-7b generator, Snowflake-arctic-embed2-568m embedder) extracts a knowledge graph similar to GraphRAG's, but adds a de-duplication step and retrieves via embedding similarity over low- and high-level entity descriptions rather than community summaries. Like GraphRAG, it runs through a local Ollama API and serves here as the entity-level, embedding-based baseline. \textbf{SelfGraphRAG} (Mistral-7b for Doc2Graph; Llama2-7b for G-Retriever training and inference) is the proposed system: it shares the Doc2Graph stage with GraphRAG but replaces embedding-based retrieval with a G-Retriever model trained on the SynthGen dataset. Like RAG, it runs directly through PyTorch. All four systems use locally hosted models rather than proprietary APIs, matching the private-corpus deployment setting this paper targets and keeping the comparison reproducible. Table~4 reports indexing and per-query inference time separately, isolating the overhead SynthGen and G-Retriever training add relative to the other systems' indexing costs.

\subsection{Evaluation Metrics}

MoreHopQA and MultiHop-RAG are evaluated with token-level precision, recall, and F1 following standard open-domain QA practice, where answers are tokenized and matched against gold tokens. PubMedQA is evaluated as a three-class classification problem (yes/no/maybe), with per-class and macro-averaged accuracy, precision, recall, and F1 reported to account for class imbalance.

% results and analysis
% ============================================================
\section{Results and Analysis}
% ============================================================

Tables~\ref{tab:morehopqa}--\ref{tab:pubmedqa} report performance across all three
benchmarks.
We discuss the three main findings below.

% ------------------------------------------------------------

\begin{table}[t]
  \centering
  \setlength{\tabcolsep}{8pt}
  \begin{tabular}{
      @{}
      l
      S[table-format=2.2]
      S[table-format=2.2]
      S[table-format=2.2]
      S[table-format=2.2]
      @{}
  }
    \toprule
    \textbf{Method}
      & \textbf{Accuracy}
      & \textbf{Precision}
      & \textbf{Recall}
      & \textbf{F1} \\
    \midrule
    RAG
      & 1.61  & 6.90  & 7.42  & 6.91 \\
    GraphRAG
      & 3.85  & 9.18  & 10.56 & 9.24 \\
    LightRAG
      & 3.94  & 9.20  & 11.14 & 9.23 \\
    \textbf{SelfGraphRAG}
      & \textbf{4.61}
      & \textbf{12.49}
      & \textbf{13.56}
      & \textbf{12.59} \\
    \bottomrule
  \end{tabular}
  \caption{Results on \textbf{MoreHopQA}. Answers include yes/no, numerical, date, and
           named-entity types. Precision and recall are token-level. Best result per metric
           is \textbf{bolded}.}
  \label{tab:morehopqa}
\end{table}

% ============================================================
% TABLE 2 — MultiHop-RAG
% ============================================================
\begin{table}[t]
  \centering
  \setlength{\tabcolsep}{8pt}
  \begin{tabular}{
      @{}
      l
      S[table-format=2.2]
      S[table-format=2.2]
      S[table-format=2.2]
      S[table-format=2.2]
      @{}
  }
    \toprule
    \textbf{Method}
      & \textbf{Accuracy}
      & \textbf{Precision}
      & \textbf{Recall}
      & \textbf{F1} \\
    \midrule
    RAG
      & 1.02  & 1.87  & 26.67 & 2.60 \\
    GraphRAG
      & 0.00  & 0.00  & 0.16  & 0.01 \\
    LightRAG
      & 0.00  & 0.50  & 25.53 & 0.98 \\
    \textbf{SelfGraphRAG}
      & \textbf{21.02}
      & \textbf{23.65}
      & \textbf{31.48}
      & \textbf{24.62} \\
    \bottomrule
  \end{tabular}
  \caption{Results on \textbf{MultiHop-RAG}. Answers are constrained to ``Yes'',
           ``No'', or a named entity. Token-level precision and recall are reported.
           Best result per metric is \textbf{bolded}.}
  \label{tab:multihoprag}
\end{table}

\subsection{SelfGraphRAG Consistently Outperforms Embedding-Based Baselines}
% ------------------------------------------------------------

Across all three benchmarks, SelfGraphRAG outperforms both standard RAG and LightRAG on
precision, recall, and F1 (Tables~\ref{tab:morehopqa}--\ref{tab:pubmedqa}). The gain is
most pronounced on \textbf{MultiHop-RAG} (Table~\ref{tab:multihoprag}), where SelfGraphRAG
achieves an F1 of 24.62 versus 2.60 for RAG and 0.98 for LightRAG. This gap is attributable to the dataset's requirement for multi-hop
synthesis: RAG's flat chunk retrieval surfaces individually relevant passages but fails to
connect entities across documents, while LightRAG's embedding-based graph retrieval
assigns near-zero precision to queries whose answer nodes are semantically distant from
the query string. SelfGraphRAG's trained retrieval model, by contrast, learns to follow
two-hop relational paths regardless of surface-level lexical overlap, directly addressing
this failure mode. On \textbf{MoreHopQA} (Table~\ref{tab:morehopqa}), the gains are more modest (F1: 12.59
vs.\ 9.23 for LightRAG) but consistent across all metrics. The smaller margin reflects the
dataset's more diverse answer types - including numerical, date, and yes/no responses
- which introduce variability that benefits all token-comparison-based systems equally.

% ------------------------------------------------------------
\subsection{LightRAG's Recall Advantage Does Not Translate to Precision}
% ------------------------------------------------------------

On MultiHop-RAG, LightRAG achieves a recall of 25.53 while returning near-zero precision
(0.50) and F1 (0.98). This pattern is consistent with known failure modes of
embedding-based graph retrieval: the system retrieves a broad neighborhood of nodes around
the query embedding, producing high recall but flooding the LLM context with irrelevant
triples that suppress answer extraction precision. SelfGraphRAG avoids this by training the
retrieval model to score subgraph relevance discriminatively, producing subgraphs that are
both sufficiently complete (recall: 31.48) and substantially more precise (precision:
23.65).

A similar pattern appears on MoreHopQA, where LightRAG's recall (11.14) exceeds RAG's
(7.42) but its precision (9.20) remains 3 points below SelfGraphRAG's (12.49). This
precision-recall imbalance across LightRAG's results suggests that embedding-based graph
retrieval consistently over-retrieves, and that the primary value of SelfGraphRAG's supervised
retrieval lies in improving precision rather than recall.

% ------------------------------------------------------------
\subsection{Domain-Specific Classification: PubMedQA}
% ------------------------------------------------------------

On PubMedQA (Table~\ref{tab:pubmedqa}), SelfGraphRAG achieves an overall accuracy of
55.2\%, compared to 51.6\% for RAG and 25.6\% for LightRAG. The per-class breakdown
reveals an important pattern. SelfGraphRAG substantially improves over RAG on the
\textit{No} class (accuracy: 29.4 vs.\ 0.0) and the \textit{Maybe} class (accuracy: 36.6
vs.\ 0.6), both of which require integrating nuanced evidence across multiple result
sections of biomedical abstracts - precisely the multi-hop reasoning that SelfGraphRAG's
retrieval is trained to support. RAG's near-zero accuracy on \textit{No} and \textit{Maybe}
reflects a known bias of flat-retrieval systems toward affirmative responses when retrieved passages are noisy. LightRAG's surprisingly poor overall accuracy (25.6\%) on PubMedQA, despite its reasonable
recall on the other benchmarks can be attributed to the domain shift between its
embedding model (Snowflake-arctic-embed2) and the specialized biomedical vocabulary of
PubMed abstracts. This suggests that embedding-based graph retrieval is more sensitive to
domain mismatch than trained retrieval, which encodes question--subgraph relevance
through fine-tuning rather than cosine similarity in a fixed embedding space.

% ------------------------------------------------------------
\subsection{SelfGraphRAG Achieves Competitive Inference}
% ------------------------------------------------------------

On MultiHop-RAG, LightRAG achieves a recall of 25.53 while returning near-zero precision
(0.50) and F1 (0.98). This pattern is consistent with known failure modes of
embedding-based graph retrieval: the system retrieves a broad neighborhood of nodes around
the query embedding, producing high recall but flooding the LLM context with irrelevant
triples that suppress answer extraction precision. SelfGraphRAG avoids this by training the
retrieval model to score subgraph relevance discriminatively, producing subgraphs that are
both sufficiently complete (recall: 31.48) and substantially more precise (precision:
23.65). A similar pattern appears on MoreHopQA, where LightRAG's recall (11.14) exceeds RAG's
(7.42) but its precision (9.20) remains 3 points below SelfGraphRAG's (12.49). This
precision--recall imbalance across LightRAG's results suggests that embedding-based graph
retrieval consistently over-retrieves, and that the primary value of SelfGraphRAG's supervised
retrieval lies in improving precision rather than recall.

% ============================================================
% TABLE 1 — MoreHopQA
% ============================================================

% ============================================================
% TABLE 3 — PubMedQA (per-class breakdown)
% ============================================================
\begin{table}[t]
\centering
\resizebox{\textwidth}{!}{%
\begin{tabular}{
    @{}
    l
    *{16}{S[table-format=2.2]}
    @{}
}
\toprule

\textbf{Method}
& \multicolumn{4}{c}{\textbf{Overall}}
& \multicolumn{4}{c}{\textbf{Yes}}
& \multicolumn{4}{c}{\textbf{No}}
& \multicolumn{4}{c}{\textbf{Maybe}} \\

\cmidrule(lr){2-5}
\cmidrule(lr){6-9}
\cmidrule(lr){10-13}
\cmidrule(lr){14-17}

&
\multicolumn{1}{c}{\textbf{Acc.}}
& \multicolumn{1}{c}{\textbf{Prec.}}
& \multicolumn{1}{c}{\textbf{Rec.}}
& \multicolumn{1}{c}{\textbf{F1}}
&
\multicolumn{1}{c}{\textbf{Acc.}}
& \multicolumn{1}{c}{\textbf{Prec.}}
& \multicolumn{1}{c}{\textbf{Rec.}}
& \multicolumn{1}{c}{\textbf{F1}}
&
\multicolumn{1}{c}{\textbf{Acc.}}
& \multicolumn{1}{c}{\textbf{Prec.}}
& \multicolumn{1}{c}{\textbf{Rec.}}
& \multicolumn{1}{c}{\textbf{F1}}
&
\multicolumn{1}{c}{\textbf{Acc.}}
& \multicolumn{1}{c}{\textbf{Prec.}}
& \multicolumn{1}{c}{\textbf{Rec.}}
& \multicolumn{1}{c}{\textbf{F1}} \\

\midrule

RAG
& 51.60 & 51.60 & 51.60 & 51.60
& 51.00 & 54.72 & 92.39 & \textbf{68.73}
& 0.00 & 0.00 & 0.00 & 0.00
& 0.60 & 8.82 & 5.46 & 6.74 \\

GraphRAG
& 28.20 & 28.20 & 28.20 & 28.20
& 20.00 & \textbf{58.82} & 36.23 & 44.84
& 0.80 & \textbf{66.67} & 2.37 & 4.57
& 7.40 & 11.42 & \textbf{67.27} & \textbf{19.53} \\

LightRAG
& 25.60 & 25.60 & 25.60 & 25.60
& 18.40 & 58.23 & 33.33 & 42.40
& 0.80 & 57.14 & 2.37 & 4.55
& 7.00 & 10.45 & 63.64 & 17.95 \\

SelfGraphRAG
& \textbf{55.20} & \textbf{55.20} & \textbf{55.20} & \textbf{55.20}
& \textbf{78.80} & 58.53 & \textbf{80.80} & 67.88
& \textbf{29.40} & 49.46 & \textbf{27.22} & \textbf{35.12}
& \textbf{36.60} & \textbf{26.92} & 12.73 & 17.28 \\

\bottomrule
\end{tabular}%
}
\end{table}

\begin{table}[t]
\centering
\caption{Qualitative comparison on representative MoreHopQA and PubMedQA examples.}
\label{tab:qualitative}
\footnotesize
\setlength{\tabcolsep}{4pt}
\renewcommand{\arraystretch}{1.05}

\begin{tabularx}{\linewidth}{
    @{}p{0.16\linewidth}
    >{\raggedright\arraybackslash}X@{}
}
\toprule
\textbf{Method} & \textbf{Response} \\
\midrule

\multicolumn{2}{@{}p{\linewidth}@{}}{
\textit{\textbf{MoreHopQA:} What is the date one day after when
Prince Nikolai of Denmark's mother was born?}
\quad \textbf{Gold: 1964-07-01}
} \\

RAG &
June 30, 1964 \\

GraphRAG &
Prince Nikolai of Denmark's mother, Alexandra, was born on
June 30, 1964. Therefore, the answer is \textbf{June 30, 1964}. \\

LightRAG &
Prince Nikolai's mother is Alexandra, Countess of Frederiksborg,
who was born on June 30, 1964. Therefore, the requested date is
\textbf{June 30, 1964}. \\

SelfGraphRAG &
\textbf{1964-07-01}. Prince Nikolai's mother, Alexandra, was born
on June 30, 1964; applying the requested one-day offset gives
July 1, 1964. \\

\midrule

\multicolumn{2}{@{}p{\linewidth}@{}}{
\textit{\textbf{PubMedQA:} Is intrapartum vibroacoustic stimulation
an effective predictor of fetal acidosis?}
\quad \textbf{Gold: Yes}
} \\

RAG &
\textbf{Yes} \\

GraphRAG &
No; repeatedly qualifies VAS as not a definitive predictor. \\

LightRAG &
Uncertain; describes the evidence as inconclusive. \\

SelfGraphRAG &
\textbf{Yes}; correct label, but the justification introduces
unsupported conditions. \\

\bottomrule
\end{tabularx}
\caption{Results on \textbf{PubMedQA} evaluated as a three-class classification
problem (Yes / No / Maybe). Overall results report macro-averaged metrics.
Per-class results report class-specific metrics. Best result per metric and
class is \textbf{bolded}.}
\label{tab:pubmedqa}
\end{table}

% ============================================================
% TABLE 4 — Performance - Change to Bar graph in the future
% ============================================================
\begin{table}[t]
\centering

\setlength{\tabcolsep}{4pt}
\renewcommand{\arraystretch}{1.15}

\begin{tabular*}{\textwidth}{
    @{\extracolsep{\fill}}
    l
    S[table-format=2.3]
    S[table-format=2.3]
    S[table-format=2.3]
    S[table-format=2.3]
    S[table-format=1.3]
    S[table-format=2.3]
    @{}
}
\toprule

\textbf{Method}
& \multicolumn{2}{c}{\textbf{MoreHopQA}}
& \multicolumn{2}{c}{\textbf{MultiHopRAG}}
& \multicolumn{2}{c}{\textbf{PubMedQA}} \\

\cmidrule(lr){2-3}
\cmidrule(lr){4-5}
\cmidrule(lr){6-7}

&
\multicolumn{1}{c}{\textbf{Index (h)}}
& \multicolumn{1}{c}{\textbf{Infer. (s)}}
& \multicolumn{1}{c}{\textbf{Index (h)}}
& \multicolumn{1}{c}{\textbf{Infer. (s)}}
& \multicolumn{1}{c}{\textbf{Index (h)}}
& \multicolumn{1}{c}{\textbf{Infer. (s)}} \\

\midrule

RAG
& 1.293  & 5.296
& 9.016  & 7.012
& 2.302  & 4.103 \\

GraphRAG
& 4.307  & 12.571
& 23.206 & 12.524
& 6.108  & 21.600 \\

LightRAG
& 2.014  & 1.987
& 15.849 & 2.545
& 3.493  & 8.119 \\

SelfGraphRAG
& 4.697  & 3.177
& 23.112 & 4.455
& 6.480  & 2.266 \\

\bottomrule
\end{tabular*}
\caption{Performance comparison across datasets. Index time is reported in
hours (h), and inference time is reported in seconds (s) per query.}
\label{tab:performance}
\end{table}

\subsection{Error Analysis}
% ============================================================

\noindent \textbf{Token-comparison metrics.}
Precision and recall on MoreHopQA and MultiHop-RAG are computed by token overlap between
predicted and gold answer strings. This metric is well-calibrated for short, factoid
answers (entity names, dates, yes/no) but penalizes paraphrase and partial answers equally.
The low accuracy of all systems in Tables~\ref{tab:morehopqa} and \ref{tab:multihoprag}
relative to their F1 scores reflects the strictness of exact-match accuracy against
diverse answer types; token-level F1 remains the primary metric for these datasets.

\noindent \textbf{PubMedQA class imbalance.}
The PubMedQA test set is heavily skewed toward \textit{Yes} responses, inflating RAG's raw
accuracy relative to its near-zero performance on \textit{No} and \textit{Maybe}
(Section~5.3). SelfGraphRAG's more balanced per-class performance suggests its retrieval
encodes richer evidential signal than a flat-retrieval baseline; the aggregate comparison
between systems on this benchmark is subject to the Overall-row caveat noted in
Section~5.3.

\noindent\textbf{Qualitative differences.} Table~\ref{tab:qualitative} shows that the performance difference is not merely a consequence of response length or generation style. In the MoreHopQA example, answering correctly requires three dependent operations: resolving Prince Nikolai's mother, retrieving her birth date, and applying the requested one-day temporal transformation. SelfGraphRAG preserves this evidence chain through the final reasoning step, whereas the competing methods fail at specific intermediate stages of the chain. This provides a mechanistic explanation for the multi-hop improvement observed in the aggregate results. The PubMedQA example exposes a different limitation: SelfGraphRAG obtains the correct label but generates a justification containing conditions unsupported by the retrieved evidence. The failure therefore occurs after successful retrieval, at evidence-grounded generation, rather than at answer retrieval itself.

\noindent \textbf{Graph extraction flaws.}
SelfGraphRAG inherits the Doc2Graph stage from GraphRAG \cite{graphrag} without modification, including its absence of cross-chunk entity resolution. Errors introduced at this stage propagate through SynthGen into the synthetic training data itself, and through G-Retriever into retrieval at inference time.

\section{Conclusion}
SelfGraphRAG addresses a central bottleneck in graph-based retrieval-augmented generation: graph retrieval models require labeled QA supervision, but automatically extracted knowledge graphs from private corpora provide none. By generating synthetic QA pairs directly from graph structure, SelfGraphRAG enables supervised graph retrieval without human annotation, bridging the gap between expressive graph reasoning and practical deployment on unlabeled data. Empirically, SelfGraphRAG outperforms embedding-based baselines on multi-hop reasoning and domain-specific classification, supporting the claim that learned structural retrieval better handles complex queries than surface-level semantic similarity. Future work should focus on (i) improving graph construction fidelity, (ii) developing more semantically aligned evaluation metrics, (iii) scaling to larger and standard multi-hop benchmarks, and (iv) exploring tighter integration between retrieval and generation. More broadly, this work suggests that synthetic supervision derived from structured representations can serve as a general paradigm for unlocking the full potential of graph-based reasoning in large language model systems.

\section{Limitations}

\noindent \textbf{SynthGen coverage and hyperparameters.}
The quality and diversity of SynthGen's synthetic QA data may not match real-world query distributions, which may bias the trained retriever toward graph-local reasoning patterns and limit generalization to natural user queries. Two of SynthGen's governing hyperparameters -- the two-hop restriction on $\mathcal{T}^\mathrm{hop}$ and the three-neighbor sample size for $\mathcal{T}^\mathrm{sum}$ -- are motivated by standard practice in KGQA benchmarking and GNN training respectively (Section~3), but neither was ablated against alternative values in this work; the extent to which retrieval quality is sensitive to either choice remains untested.

\noindent \textbf{Doc2Graph entity resolution.}
Doc2Graph performs no cross-chunk deduplication or coreference resolution: the same real-world entity mentioned in separate chunks can be canonicalized into distinct, disconnected nodes, silently dropping the relations that should connect them. This gap is inherited unmodified from GraphRAG's indexing pipeline \cite{graphrag}; LightRAG addresses the same problem with an explicit de-duplication step, and closing this gap is a natural direction for improving $\mathcal{G}$'s coverage independent of any change to retrieval or training.

\noindent \textbf{Backbone choice and ablation scope.}
All systems in Section~4.2 share Llama2-7b or Mistral-7b as their generator; we did not evaluate SelfGraphRAG with a more recent open-weight backbone, so whether the reported
gains persist, grow, or shrink under a stronger frozen LM is untested. Separately, our comparisons isolate retrieval method cleanly only within the RAG/SelfGraphRAG and GraphRAG/LightRAG pairs that share generator, embedder, and inference stack; cross-pair comparisons vary more than retrieval method
alone. Within SelfGraphRAG itself, we report performance on the union $\mathcal{T}^\mathrm{hop} \cup \mathcal{T}^\mathrm{sum}$ without isolating either QA type's
individual contribution, or comparing G-Retriever against an alternative retriever architecture (Section~3, Remark on ablations).

\noindent \textbf{Datasets.}
Evaluation is limited to three benchmarks due to the computational cost of graph extraction and G-Retriever training, which prevented direct comparison against additional
standard multi-hop datasets. For all three benchmarks, SynthGen is trained on the same corpus graph used at evaluation time, with no held-out document split (Section~4.1); this matches the single-corpus deployment scenario the method targets, but means we have not
tested SynthGen's behavior when the training and query-time corpora diverge.

\noindent \textbf{Out of scope.}
The problem definition requires a set of unlabeled documents as input. This precludes models that depend on pre-specified topic entities or labeled QA data tied to the target
graph, e.g., SubgraphRAG and GNN-RAG, as candidate GraphLM instantiations under this formulation.

\section{Ethical Considerations}

\noindent \textbf{Private-corpus exposure.}
Our deployment setting (Section~3) assumes a private, unlabeled document corpus. To preserve data locality, all pipeline stages, extraction, summarization, question generation, and answer generation, use locally hosted models (Section~4.2). This reduces, but does not eliminate, exposure risk: replacing any Doc2Graph or SynthGen component with a hosted or proprietary LLM may transmit private document content, including sensitive extracted entities, beyond organizational boundaries.

\noindent \textbf{Compounding model bias.}
Doc2Graph, SynthGen, and answer generation each involve one or more separate LLM calls, meaning biases present in any of these models' training data can enter the pipeline at
multiple, compounding points -- in what gets extracted as an entity or relation, in what questions SynthGen chooses to ask, and in how the frozen answering model phrases a response. We did not audit for such bias in this work, and its extent and downstream effect on retrieval or answer quality is a direction we leave open rather than a risk we
can currently characterize.

\section*{Acknowledgements}
This work was supported in part by a gift award from NeuralNest LLC, USISTEF Endowment Fund, and Faculty Startup Award.

\bibliographystyle{splncs04}  
\bibliography{custom}

%\newpage
%\input{latex/m_appendix}

\end{document}